\documentclass[preprint,12pt,authoryear]{elsarticle}

\usepackage{amssymb}
\usepackage{subfigure}
\usepackage{multirow}
\usepackage{xcolor}
\usepackage{amsmath}
\usepackage{booktabs}
\usepackage[colorlinks,linkcolor=blue,anchorcolor=blue,citecolor=blue,urlcolor=blue]{hyperref}

\journal{IPM}

\begin{document}

\begin{frontmatter}

\title{Beyond Textual Knowledge: Leveraging Multimodal Knowledge Bases for Enhancing Vision-and-Language Navigation}

\address[label1]{School of Computer Science and Technology, Xinjiang University, Urumqi, China}
\address[label2]{Joint International Research Laboratory of Silk Road Multilingual Cognitive Computing}
\cortext[cor1]{Corresponding author}

\author[label1,label2]{Dongsheng Yang}
\ead{107552304180@stu.xju.edu.cn} 

\author[label1,label2]{Yinfeng Yu\corref{cor1}}
\ead{yuyinfeng@xju.edu.cn}

\author[label1,label2]{Liejun Wang}
\ead{wljxju@xju.edu.cn}

\begin{abstract}

Vision-and-Language Navigation (VLN) requires an agent to navigate through complex unseen environments based on natural language instructions. However, existing methods often struggle to effectively capture key semantic cues and accurately align them with visual observations. To address this limitation, we propose Beyond Textual Knowledge (BTK), a VLN framework that synergistically integrates environment-specific textual knowledge with generative image knowledge bases. BTK employs Qwen3-4B to extract goal-related phrases and utilizes Flux-Schnell to construct two large-scale image knowledge bases: R2R\_GP and REVERIE\_GP. Additionally, we leverage BLIP-2 to construct a large-scale textual knowledge base derived from panoramic views, providing environment-specific semantic cues. These multimodal knowledge bases are effectively integrated via the Goal-Aware Augmentor and Knowledge Augmentor, significantly enhancing semantic grounding and cross-modal alignment. Extensive experiments on the R2R dataset with 7,189 trajectories and the REVERIE dataset with 21,702 instructions demonstrate that BTK significantly outperforms existing baselines. On the test unseen splits of R2R and REVERIE, SR increased by 5\% and 2.07\% respectively, and SPL increased by 4\% and 3.69\% respectively. The source code is available at
https://github.com/yds3/IPM\_BTK/.

\end{abstract}

\begin{graphicalabstract}
	\includegraphics[width=32em]{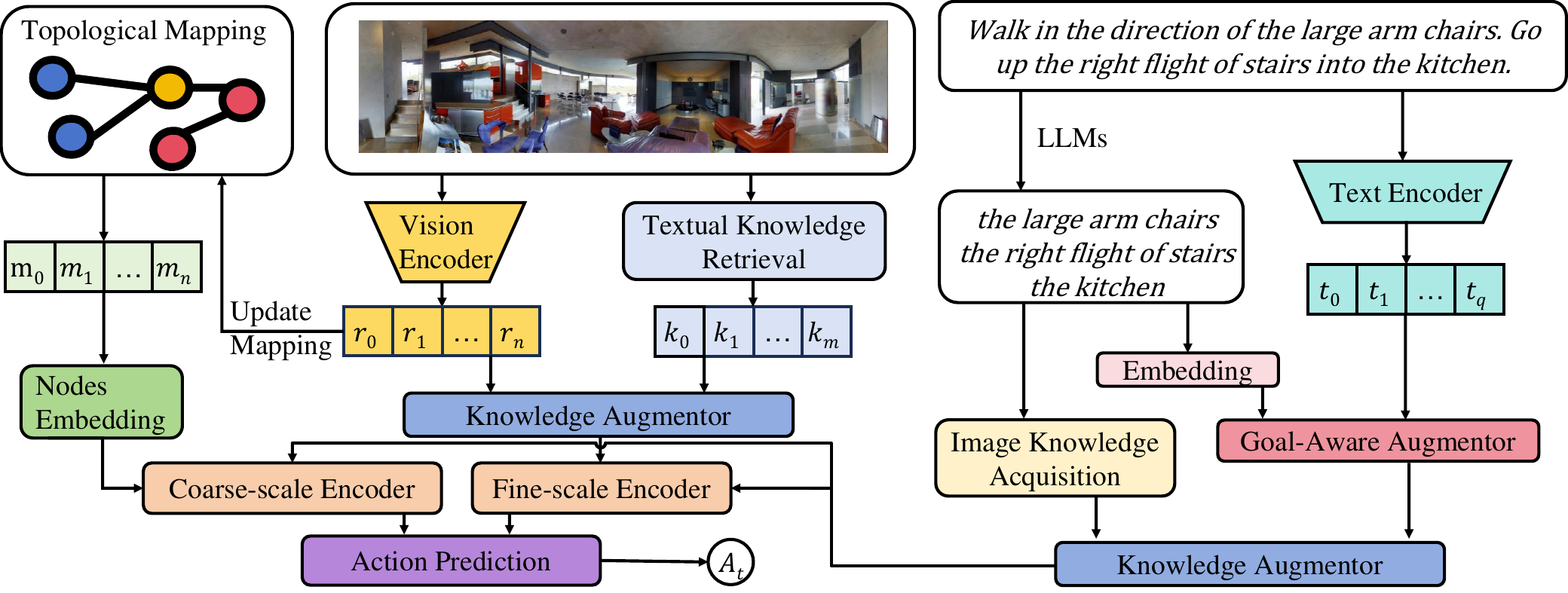}
\end{graphicalabstract}

\begin{highlights}
\item Proposed a VLN framework that integrates environment-specific textual knowledge with generative visual knowledge bases.

\item Constructed R2R\_GP and REVERIE\_GP, two large-scale image knowledge bases for VLN.

\item Introduced a goal-aware augmentor to enhance target perception and alignment.
\end{highlights}


\end{frontmatter}

\section{Introduction}

In recent years, the rapid progress of embodied intelligence \citep{huang2026orchestrating}and multimodal understanding has brought increasing attention to language guided perceptual tasks, where agents must interpret linguistic instructions and interact with visual environments. Among these, VLN \citep{anderson2018vision,han2025roomtour3d,huang2025action} stands out as a fundamental benchmark, requiring an agent to navigate through unseen 3D environments solely based on natural language commands. Solving this task demands not only accurate semantic comprehension of the instructions but also dynamic perception, cross-modal reasoning, and sequential decision-making.

To facilitate research in this area, several benchmarks have been introduced, such as R2R \citep{anderson2018vision} which focuses on fine-grained step-by-step navigation and REVERIE \citep{qi2020reverie}, which emphasizes goal-oriented reasoning and object grounding in more realistic scenarios.Despite the remarkable progress achieved by recent Transformer-based approaches \citep{lin2022adapt,wang2024lookahead,gao2023adaptive}, several critical challenges in VLN remain unresolved.
First, most existing methods treat the entire instruction as a single sequence, which makes it difficult to effectively identify and emphasize key semantic elements such as target objects or spatial relationships. Second, many approaches rely on abstract and general commonsense knowledge, which often fails to capture the specific context of the navigation environment, leaving a significant “semantic gap” between abstract concepts and concrete visual instances. Moreover, most existing methods are limited to purely textual knowledge, which does not address the core cross-modal challenges. As a result, the alignment between linguistic goals and visual observations remains shallow and fragile, especially in cluttered or occluded environments.
\begin{figure*}[t]
\centering
\includegraphics[width=5in]{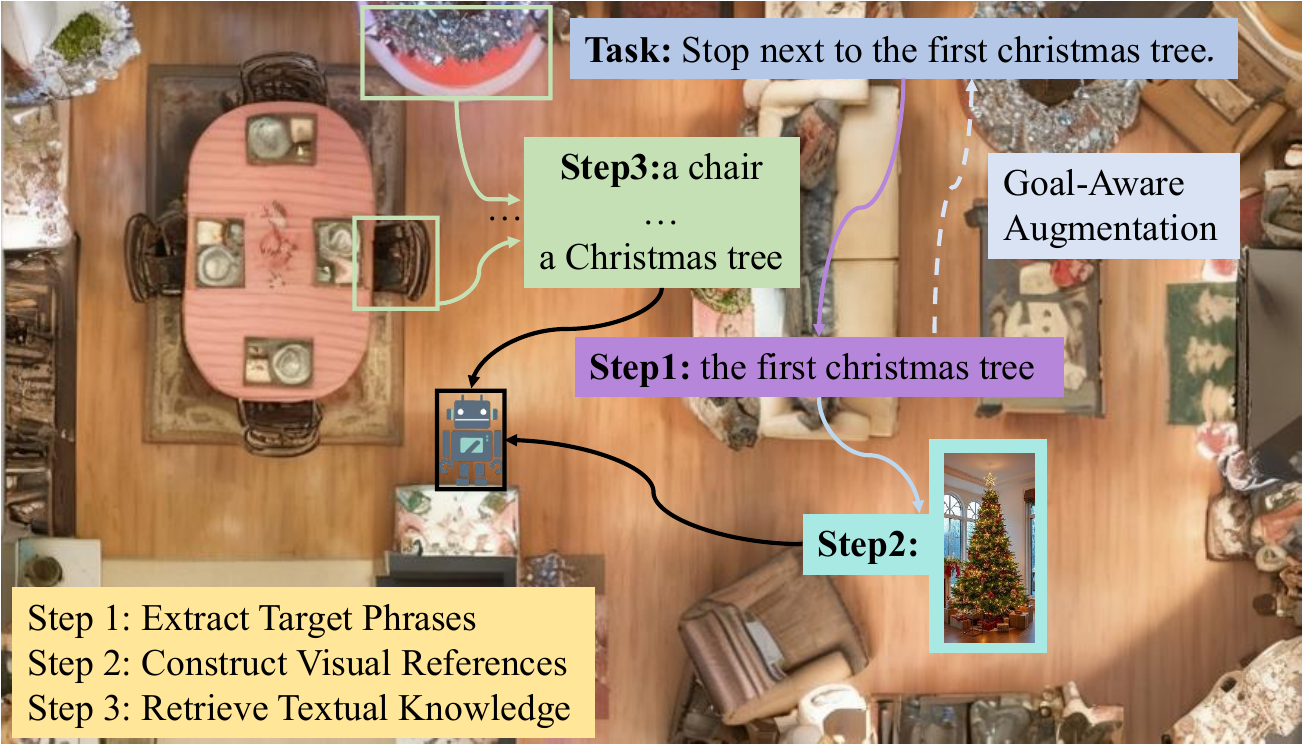}
\caption{ BTK mainly consists of three stages: (1) extracting goal-related phrases using Qwen3-4B and enhancing their semantics through a goal-aware augmentation module; (2) generating visual exemplars to construct the image knowledge base; and (3) retrieving complementary textual knowledge from BLIP-2 to achieve robust cross-modal alignment.}
\label{top}
\end{figure*}

We propose BTK, a unified framework that enhances instruction comprehension and environmental grounding through goal-aware augmentation and multimodal knowledge modeling , aiming to address the challenges of coarse-grained instruction understanding , fragile cross-modal alignment , and the 'semantic gap' resulting from existing methods' reliance solely on external textual commonsense in VLN. We define 'knowledge' as supplementary information that is not explicitly provided in the instruction or current view but is crucial for semantic understanding and contextual reasoning. We leverage this knowledge in two forms: (1) Textual Knowledge, which refers to fine-grained semantic descriptions of the environment to provide contextual cues for ambiguous visual scenes; and (2) Image Knowledge, which refers to concrete visual exemplars corresponding to key target phrases in the instruction, providing direct references to bridge the language-vision gap.

 As illustrated in  \autoref{top}, BTK processes natural language instructions and visual observations in three stages. First, for a given instruction, the large language model is used to extract goal-related phrases as semantic anchors, which are further emphasized by a goal-aware augmentation module to guide the agent’s focus toward informative navigational cues. Second, based on the definition of image knowledge, a text-to-image generation model converts these target phrases into visual exemplars, constructing two large-scale image knowledge bases, R2R\_GP and REVERIE\_GP, to provide explicit visual grounding for the instruction targets. Finally, textual knowledge is retrieved from a text knowledge base generated using BLIP-2 to effectively compensate for visual information missing due to cluttered or occluded environments. Extensive experiments on the R2R and REVERIE benchmarks demonstrate that BTK significantly outperforms strong baselines across multiple metrics, achieving state-of-the-art performance. The main contributions of this work are as follows:
\begin{enumerate}
\item 
We propose BTK, a VLN framework leveraging generative multimodal knowledge bases. It uniquely combines environment-specific textual knowledge with generative image knowledge produced by diffusion models, addressing the limitations of prior text-only methods and achieves superior performance on the R2R and REVERIE benchmarks.
\item We construct two large-scale image knowledge bases, namely R2R\_GP and REVERIE\_GP, which provide visual references for goal phrases and effectively enhance cross-modal alignment between language and vision.
\item We introduce a Goal-Aware Augmentor that leverages an LLM to extract complete, descriptive goal phrases, surpassing the limitations of prior semantic extractors that rely on isolated or abstract nouns.

\end{enumerate}

\section{Related Work}

\subsection{Vision-and-Language Navigation}

VLN \citep{anderson2018vision} is a challenging task in embodied AI with broad real-world applications. Since its introduction, VLN has attracted substantial attention from the research community \citep{wang2018look,jain2019stay,lin2023learning,he2023frequency}. Unlike visual question answering, which performs one-shot inference on static images or short videos \citep{qiu2024explainable}, VLN requires an agent to follow natural language instructions to navigate in unfamiliar and complex 3D environments through sequential decision-making and interactive control, demanding a deep understanding of linguistic semantics and effective alignment among vision, language, and action.

Early studies focused on dataset construction and training paradigms. Datasets such as R2R \citep{anderson2018vision}, REVERIE \citep{qi2020reverie}, SOON \citep{zhu2021soon}, and Room-Across-Room \citep{ku2020room} laid a solid foundation for VLN research. To alleviate data scarcity, various augmentation techniques have been proposed \citep{li2022envedit,wang2023res,wang2023scaling}, aiming to improve training efficiency and generalization. For example, PANOGEN \citep{li2023panogen} utilizes image captions and text-to-image diffusion models to generate structurally grounded virtual environments, while ORIST \citep{qi2021road} fuses panoramic and object-level features through parallel encoders to enhance visual representation.

With the rise of cross-modal pretraining, more recent approaches have leveraged large-scale pretrained models for better language-vision alignment. DUET \citep{chen2022think} adopts a pretraining-finetuning paradigm and designs a dual-scale Transformer to improve cross-modal understanding. HOP+ \citep{qiao2023hop+} introduces history-aware modeling and a memory-based architecture to resolve discrepancies between pretraining and downstream tasks. GOAT \citep{wang2024vision} incorporates causal inference mechanisms into VLN, aiming to mitigate the impact of data bias on agent performance.

Furthermore, recent research has begun to focus on leveraging finer-grained navigation sub-goals and explicit 3D scene representations to enhance navigation performance. For instance, TD-STP\citep{zhao2022target} introduces an imagined scene tokenization mechanism to explicitly estimate long-term navigation goals. Meanwhile, a series of studies have been dedicated to enhancing the agent's spatial awareness by constructing stronger 3D environment representations\citep{wang2025dynam3d,wang2025g3d,gao20253d}, including Bird's-Eye-View (BEV) scene graphs\citep{liu2023bird} and volumetric environment representations\citep{liu2024volumetric}.

\subsection{Semantic Information in VLN}

Semantic information within language instructions plays a vital role in enhancing VLN performance by guiding the agent’s attention toward discriminative environmental details. SOAT \citep{moudgil2021soat} jointly models scene-level and object-level features, significantly improving navigational precision. OAAM \citep{qi2020object} employs attention mechanisms to extract semantic representations of objects and actions from instructions, thereby enhancing instruction comprehension.

In contrast, DSRG \citep{Wang2023DSRG} utilizes the Spacy toolkit to extract nouns and verbs from instructions for semantic augmentation. However, this method faces two key limitations: (1) the extracted terms often include abstract nouns (e.g., “left”), which provide ambiguous visual grounding, and (2) it tends to overlook descriptive modifiers in noun phrases (e.g., “red” in “red chair”), reducing the accuracy of target localization.

While these methods validate the utility of semantic information, they often focus on extracting isolated semantic units (nouns, verbs), which can fragment the instruction's core intent. To address these issues, we incorporate large language models (LLMs) with superior contextual reasoning ability. However, our approach is not merely a tool-level substitution (i.e., replacing Spacy with an LLM). Instead, we introduce a conceptual shift: we leverage LLMs to extract complete, descriptive goal phrases (e.g., "red chair") rather than isolated or abstract nouns (e.g., "chair," "left"). This transition from 'isolated terms' to 'groundable phrases' preserves critical modifier information and captures the full semantic intent necessary for detailed instruction parsing. Furthermore, these complete phrases serve as the foundation for our generative image knowledge base, a component not present in prior semantic enhancement methods.

\subsection{Knowledge Bases in VLN}

Recent studies have shown that external knowledge can significantly improve multimodal systems’ performance on understanding and reasoning tasks\citep{gao2024enhancing,gao2025visual}. Common sources include structured knowledge graphs (e.g., DBpedia \citep{auer2007dbpedia}) and conceptual semantic networks (e.g., ConceptNet \citep{liu2004conceptnet}). In VLN, several methods have incorporated external knowledge to enhance the agent’s reasoning capacity. CKR \citep{gao2021room} utilizes commonsense knowledge from ConceptNet to model semantic relationships between rooms and objects. KERM \citep{li2023kerm} retrieves relevant knowledge from Visual Genome \citep{krishna2017visual} to enrich panoramic vision features. ACK \citep{mohammadi2024augmented} also retrieves entity-level knowledge from ConceptNet to assist the agent in perception and decision-making.

The aforementioned methods, while beneficial, are constrained by two fundamental limitations: (1) They are exclusively text-only knowledge approaches, retrieving textual facts or relations. This fails to directly bridge the core cross-modal gap between abstract language and concrete visual appearance. (2) They rely on general-purpose knowledge bases that are not tailored to the specific visual and structural properties of indoor navigation environments.

Our BTK framework differs fundamentally from these approaches in both aspects. First, we introduce a multimodal knowledge resource, moving beyond the text-only limitation. We propose two large-scale image knowledge bases that translate semantic goal phrases from instructions into concrete visual exemplars. Second, our textual knowledge base is environment-specific. Instead of using general commonsense, we generate it directly from the panoramic views of the navigation environments using BLIP-2. This dual-knowledge approach, combining environment-specific textual knowledge with generative image knowledge, facilitates stronger alignment between language understanding and visual perception, thereby enhancing the agent’s cross-modal modeling capacity.

\section{Preliminaries}

\subsection{Problem Formulation}

Our model focuses on the VLN task in a discrete indoor environment. The environment is modeled as an undirected graph $G = \{V, E\}$, where the node set $V = \{V_i\}_{i=1}^{k}$ denotes $k$ navigable locations, and the edge set $E$ defines connectivity between them. In datasets like R2R \citep{anderson2018vision} and REVERIE \citep{qi2020reverie}, at the initial timestep $t_0$, the agent is randomly placed in a complex, unfamiliar scene and receives a natural language instruction.

The agent is equipped with a GPS sensor and an RGB camera for position tracking and panoramic vision acquisition. A navigation instruction is denoted by $I = \{w_i\}_{i=1}^{L}$, where $L$ is the length of the instruction and $w_i$ represents the embedded word vectors. The panoramic image at time $t$ consists of 36 directional views $R_t = \{r_i\}_{i=1}^{36}$, where each $r_i$ contains a feature vector and its corresponding direction $(\theta_{1,i}, \theta_{2,i})$, denoting azimuth and elevation angles.

During navigation, the agent selects the next action from a discrete action space based on its current multimodal state to move to a neighboring node. The process continues until the agent decides to stop or reaches the maximum number of allowed steps. After termination, the agent predicts the final location of the target object. A navigation attempt is deemed successful if the Euclidean distance between the predicted and ground-truth target is within 3 meters.

\subsection{Large Models}

Thanks to large-scale pretraining on internet corpora, large pretrained models exhibit strong generalization and linguistic capabilities, and have achieved remarkable performance across various tasks. In our proposed BTK framework, we integrate four powerful pretrained models: Qwen3-4B \citep{yang2025qwen3}, BLIP-2 \citep{li2023blip}, CLIP \citep{radford2021learning}, and Flux-Schnell, for target recognition, vision-language alignment, and image generation.

Qwen3-4B is a lightweight yet capable language model known for its context-aware reasoning. It performs well in tasks like target phrase extraction with minimal prompting.

BLIP-2 is a vision-language pretrained model employing a two-stage pretraining strategy. It connects a vision encoder (ViT) with a large frozen language model via a lightweight Q-Former module, enabling efficient image-text alignment.

CLIP is a vision-language model trained on 400 million image-text pairs using contrastive learning. With ViT or ResNet as the image encoder and Transformer as the text encoder, CLIP learns unified representations for images and texts, enabling strong zero-shot recognition and retrieval capabilities.

Flux-Schnell is a powerful text-to-image diffusion model with a 12B-parameter parallel transformer architecture. It can generate high-quality images in 1–4 steps and demonstrates outstanding prompt comprehension and generation efficiency, making it suitable for local deployment.

\begin{figure*}[t]
\centering
\includegraphics[width=5in]{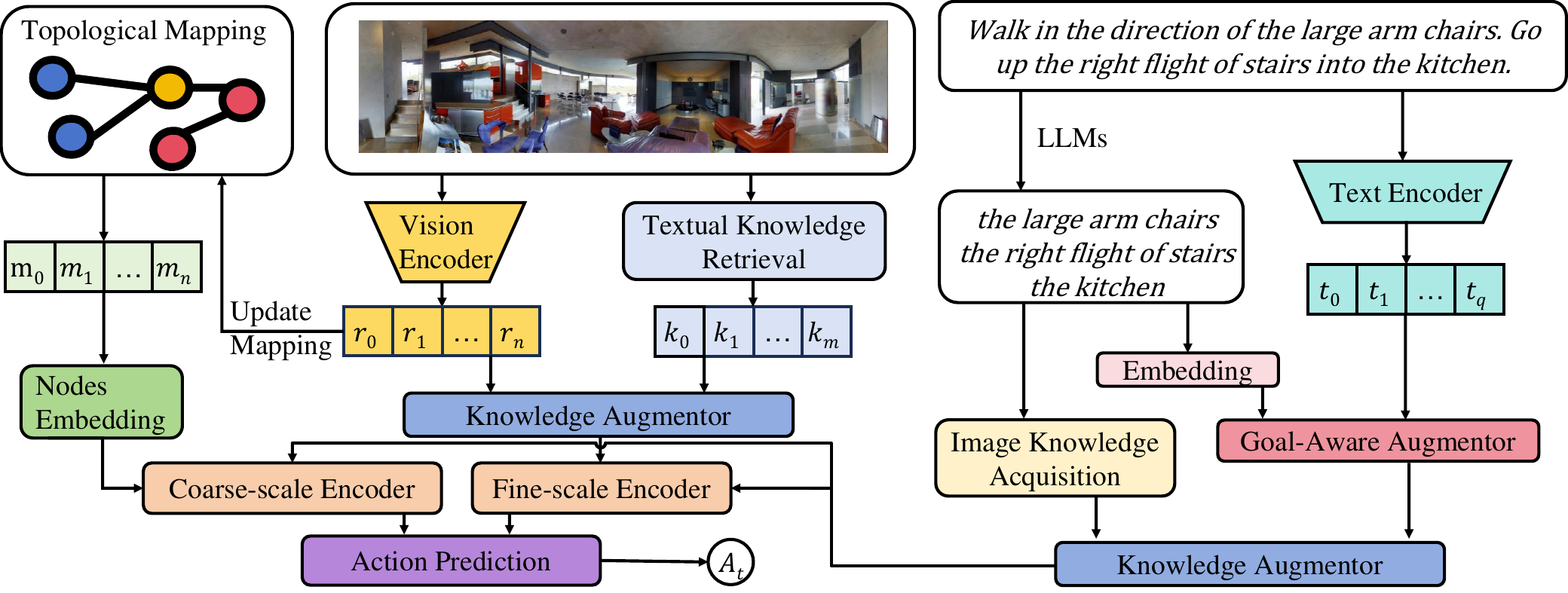}
\caption{Overview of the BTK. 
The model first extracts sub-goals from the natural language instruction. Subsequently, the Goal-Aware Augmentor utilizes these sub-goals to simultaneously reinforce the instruction's semantics and generate corresponding image knowledge. Concurrently, the model acquires textual knowledge by matching visual information with a textual knowledge base. Finally, the Knowledge Augmentor integrates this multimodal knowledge to jointly enhance the instruction and visual inputs.}
\label{kjt}
\end{figure*}

\section{Method}
\label{sec:method}

We adopt DUET as our baseline. DUET takes three primary inputs: a dynamically constructed topological map, natural language instructions, and the agent's current panoramic view. The topological map consists of three types of nodes: visited nodes, the current node, and navigable nodes. At each step, DUET feeds these features into a dual-scale graph Transformer, composed of a fine-scale encoder and a coarse-scale encoder, to predict the agent’s next action.

We propose a novel framework for VLN called BTK. This framework is designed to help agents leverage multimodal knowledge for goal recognition and path planning in complex environments, as illustrated in  \autoref{kjt}. The core idea of BTK is as follows: it first employs a large language model (LLM) to extract sub-goals and enhances the semantic information of instructions through a goal-aware enhancer; simultaneously, it uses Flux-Schnell to generate corresponding image-based knowledge. The framework then integrates the pre-constructed text and image knowledge bases to achieve more precise cross-modal alignment and decision-making.

To this end, BTK introduces two key modules: the Goal-Aware Augmentor and the Knowledge Augmentor. The former extracts goal-related semantics from instructions and generates more focused language representations, while the latter leverages both visual and textual knowledge bases to improve the model's perception and reasoning in complex scenes.

In VLN tasks, the agent must plan paths and recognize goals based on natural language instructions. However, instructions in real-world scenarios often contain redundancy, ambiguity, or vague descriptions, making it difficult for the agent to accurately identify navigation targets. To address this issue, we employ a large language model to structurally analyze instructions, explicitly extracting goal-related phrases as semantic anchors. These subgoals are then fused with the original instruction via the Goal-Aware Augmentor, resulting in enhanced instructions that are more structured and goal-centric.

\subsection{Subgoal Extraction}
When extracting goal-related phrases from instructions, we consider three key principles: (1) extract all explicitly mentioned goal anchors; (2) preserve attribute-modified descriptions (e.g., “red sofa”); and (3) avoid introducing out-of-instruction content. To achieve this, we utilize the Qwen3-4B language model with the following customized prompt: Extract multiple detailed noun phrases from the following instruction, list them numerically, each noun phrase must be written in English and end with a period, do not output other information.

This prompt enables efficient and accurate subgoal extraction, denoted as $I_g$. On average, each instruction in the R2R dataset yields 4.3 subgoals, while REVERIE instructions yield 2.3.

\subsection{Goal-Aware Enhancement}
\begin{figure}[!t]
\centering
\includegraphics[width=5in]{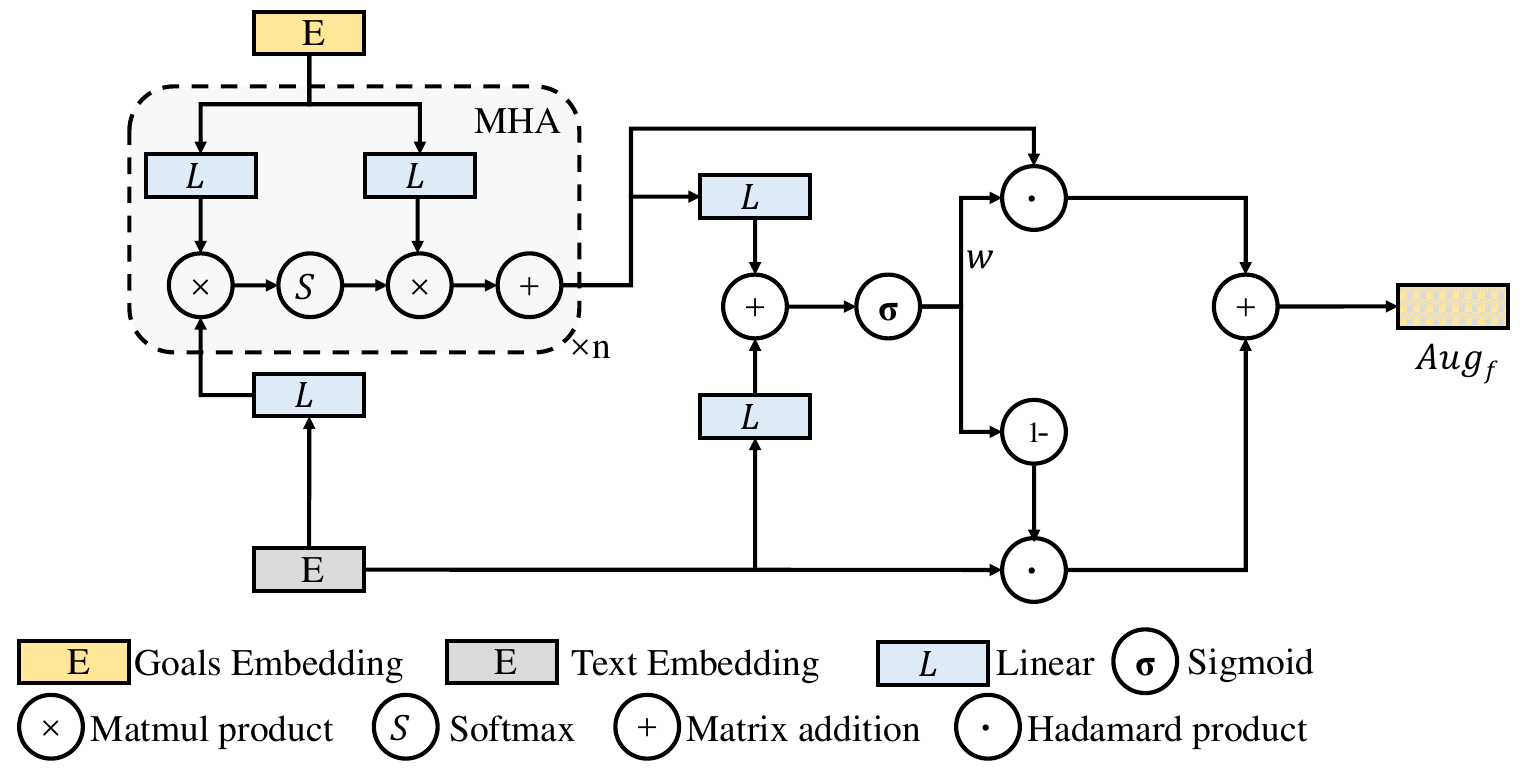}
\caption{Architecture of the Goal-Aware Augmentor.}
\label{gaa}
\end{figure}

\subsubsection{Goal-Aware Augmentation}

The extracted subgoals are sequentially combined with the original instruction and fed into the Goal-Aware Augmentor, as shown in \autoref{gaa}. This module primarily consists of a Multi-Head Attention (MHA) mechanism. Each token in the subgoals is embedded into a 768-dimensional space. Meanwhile, the original instruction $I = \{w_i\}_{i=1}^{L}$ is encoded via BERT to produce features $T$. The instruction features are used as queries (Q), and subgoal embeddings serve as keys (K) and values (V) in the MHA module to yield enhanced features $T^{'}$.

To balance enhancement and original semantics, we introduce a dynamic fusion mechanism. Specifically, a Sigmoid function computes the fusion weights $\omega$, which are used to combine the enhanced features $T^{'}$ with the original features $T$, yielding the final instruction representation $T^{''}$:
\begin{align}
(\mathcal{I},  \mathcal{I}_g^{'}) &= \text{Embedding}({I}, {I}_g) , \\
T &= \text{BERT}(\mathcal{I}), \label{eq2} \\
T^{'} &= \text{MHA}(T,\mathcal{I}_g^{'}), \label{eq3} \\
\omega &= \text{Sigmoid} (T W_g + T^{'} W_c + b_I), \label{eq4} \\
T^{''}&= \omega \odot  T^{'} + (1 - \omega) \odot T. \label{eq5}
\end{align}
where $W_g$ and $W_c$ are learnable parameters, and $b_I$ is a learnable bias term for the instruction fusion gate.

\subsection{Knowledge Base Construction}
\label{subsec:tse}

In VLN, external knowledge bases provide rich semantic information that aids the agent in understanding complex environments. To enable comprehensive multimodal enhancement, we construct two types of knowledge bases: a textual knowledge base and an image knowledge base. The former provides semantic complementarity to visual content, while the latter facilitates fine-grained alignment between textual and visual features.

\subsubsection{Textual Knowledge Base}

To construct a high-quality textual knowledge base for indoor scenes, we utilize the pre-trained vision-language model BLIP-2 to extract descriptive information from panoramic images in the Matterport3D environment. Each panorama is divided into 36 sub-view images. We apply the prompt “a photo of” to each view to guide BLIP-2 in generating textual descriptions of room layouts and object relations, e.g., “a bathroom with two sinks and a shower.” We collect approximately 380K textual knowledge entries, covering diverse indoor scenarios and common object combinations.
\begin{figure}[!t]
\centering
\includegraphics[width=3.35in]{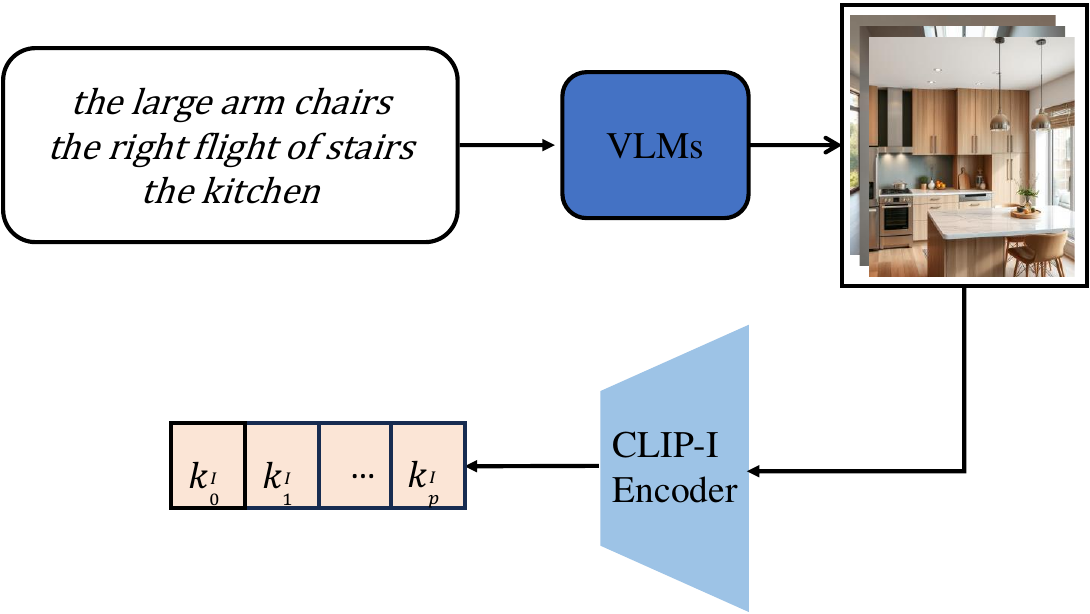}
\caption{Architecture of image knowledge acquisition.}
\label{ika}
\end{figure}

To enhance subgoal-scene alignment, we build an image knowledge base by generating visually consistent and semantically relevant images. Given a subgoal, we compose prompts such as “[subgoal] of the indoor environment in real estate” and input them to the Flux-Schnell diffusion model, which is capable of generating high-quality indoor imagery.

% On the R2R dataset, we generate approximately 93K image-based knowledge entries (about 4.3 per instruction), forming the R2R\_GP image knowledge bases. On the REVERIE dataset, we generate approximately 50K image-based knowledge entries (about 2.3 per instruction), forming the REVERIE\_GP image knowledge bases.

On the R2R dataset, we generate approximately 93K image-based knowledge entries, with an average of 4.3 entries per instruction, forming the R2R\_GP image knowledge bases. On the REVERIE dataset, we generate approximately 50K image-based knowledge entries, with an average of 2.3 entries per instruction, forming the REVERIE\_GP image knowledge bases.

\subsection{Knowledge Enhancement}
\label{subsubsec:oap}

\subsubsection{Textual Knowledge Retrieval}

Panoramic images contain fine-grained information such as color, small objects, and spatial layout. However, relying on a single textual knowledge entry often fails to fully capture these elements and tends to introduce subjectivity. To address this issue, we encode the 36 sub-views and textual knowledge entries using CLIP’s visual and textual encoders, respectively. We then compute their similarity via dot product, using cosine similarity, and retrieve the top-5 most relevant textual entries for each sub-view as its textual knowledge set ($K = \{{k_i}\}_{i=1}^{n}$.

\subsubsection{Image Knowledge Acquisition}

As shown in  \autoref{ika}, for each subgoal, we generate corresponding images using customized prompts and the Flux-Schnell model. Notably, to efficiently save time and reduce unnecessary resource consumption, we do not generate images and run the model simultaneously. Instead, we first generate the images to construct the image knowledge base, extract features from the image knowledge base using CLIP-B/16~\citep{radford2021learning}, and then index the image knowledge features 
$K' = \{{k_i^I}\}_{i=1}^{p}$
according to the corresponding instructions.

\subsubsection{Knowledge Augmentor}

Instead of naïvely concatenating knowledge with inputs, we design a Knowledge Augmentor to selectively fuse the most relevant multimodal knowledge with the current task context.The Knowledge Augmentor is a reusable fusion module that plays a crucial role in our framework, serving to fuse image knowledge with instruction features and textual knowledge with visual features.

\begin{figure}[!t]
\centering
\includegraphics[width=3.35in]{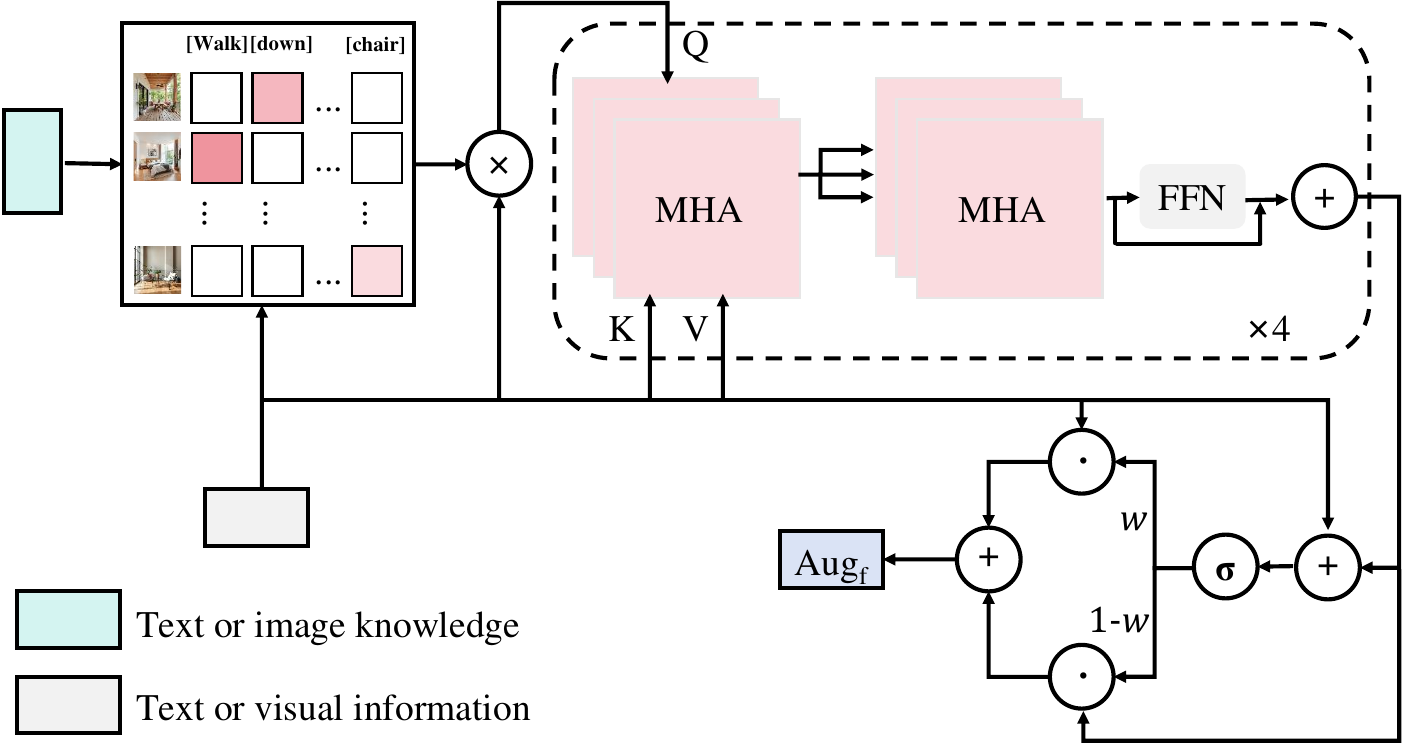}
\caption{Architecture of the Knowledge Augmentor.}
\label{ka}
\end{figure}

As shown in  \autoref{ka}, we first compute a correlation matrix $Mat$ between the image knowledge features $K'$ and the instruction features $T^{''}$, aiming to filter out irrelevant information between them. The computation is defined as:
\begin{align}
Mat = (K' W_k)(T^{''} W_t)^\top / \sqrt{d} .\label{eq1}
\end{align}
where $W_k$ and $W_t$ are learnable projections and $d$ is the feature dimension.

Applying Softmax yields attention weights:
\begin{equation}
W= \mathrm{softmax}(Mat) .
\end{equation}

We compute the weighted sum to obtain a condensed knowledge representation:
\begin{equation}
K^{''} = W \cdot Mat .
\end{equation}

This is used in an MHA layer to enhance the instruction:
\begin{equation}
\tilde{T} = \mathrm{MHA}(K^{''}, T^{''}).
\end{equation}

A gating mechanism fuses the original and enhanced features:
\begin{equation}
\mathrm{g} = \sigma( W_1 \tilde{T} + W_2 T^{''} +b).
\end{equation}
where $\sigma(\cdot)$ denotes the Sigmoid function, $W_1$ and $W_2$ are learnable parameters, and $b$ is a learnable bias term for the instruction fusion gate..

The final augmented instruction is:
\begin{equation}
T_{aug} = g\odot T^{''} + (1 - g) \odot \tilde{T}.
\end{equation}
where $\odot$ denotes element-wise multiplication.

For processing textual knowledge, we employ the same strategy. The textual knowledge features $K$, along with the visual features $R_t^{'}$ extracted via CLIP-B/16 \citep{radford2021learning}, are input to the Knowledge Augmentor, yielding the enhanced visual features $R_t^{''}$.

\subsection{Action Prediction}

The enhanced visual features $R_t^{''}$ and instruction features $T_{aug}$ are fed into DUET's Fine-scale Encoder to compute local action scores $S_i^f$. The topological map features $M$ and $T_{aug}$ are input into the Coarse-scale Encoder to obtain global action scores $S_i^c$. The local scores $S_i^f$ are then converted to global scores $\hat{S}_i^c$, which are fused with $S_i^c$ to yield the final action prediction probability $S_i$. The action with the highest score is selected as the next navigation step.

\section{Experiment}
\subsection{Experimental Setup}
\subsubsection{Datasets}

To validate the effectiveness of the proposed method, we conduct extensive experiments on two widely-used VLN benchmark datasets: R2R~\citep{anderson2018vision} and REVERIE~\citep{qi2020reverie}. As shown in  \autoref{tab:1}, both datasets are discrete datasets constructed based on indoor environments and are strictly split into four subsets: training, seen validation, unseen validation, and unseen test.

\begin{table}[t]
\centering

\caption{Dataset split statistics for VLN tasks.}
\label{tab:1}
\begin{tabular}{lllllllll}
\toprule
\multirow{2}{*}{Dataset} & \multicolumn{2}{l}{Train} & \multicolumn{2}{l}{Val seen} & \multicolumn{2}{l}{Val unseen} & \multicolumn{2}{l}{Test unseen} \\ 
\cmidrule(r){2-3} \cmidrule(lr){4-5} \cmidrule(lr){6-7} \cmidrule(l){8-9}
 & House & Instr & House & Instr & House & Instr & House & Instr \\ 
\midrule
R2R & 61 & 14025 & 56 & 1020 & 11 & 2349 & 18 & 4173 \\
REVERIE & 60 & 10466 & 46 & 1423 & 10 & 3521 & 16 & 6292 \\
\bottomrule
\end{tabular}
\end{table}

The R2R dataset comprises a total of 7,189 navigation trajectories, each annotated with three natural language instructions averaging 32 words in length. These instructions provide step-by-step guidance. The test set includes 18 scenes and 4,173 instructions, the unseen validation set contains 11 scenes and 2,349 instructions, and the remaining 61 scenes are divided into the training set with 14,025 instructions and the seen validation set with 1,020 instructions.

The REVERIE dataset consists of 21,702 instructions with an average length of 18 words and a vocabulary size exceeding 1,600. Unlike the step-by-step navigation instructions in R2R, REVERIE adopts goal-oriented instructions that require the agent not only to navigate to the correct location but also to identify the correct object bounding box upon stopping. Each panoramic image in REVERIE is annotated with predefined object bounding boxes to support training and evaluation of grounding capabilities.

\subsubsection{Evaluation Metrics}

To ensure fair comparisons, we adopt standard and widely-used evaluation metrics in VLN tasks. For the R2R dataset, we use the following metrics: Navigation Error (NE), which measures the average Euclidean distance (in meters) between the agent’s final stopping position and the ground truth goal; Success Rate (SR), defined as the percentage of episodes where the agent stops within 3 meters of the goal location; Oracle Success Rate (OSR), which computes SR assuming the agent stops at the closest point to the goal along its trajectory; and Success weighted by Path Length (SPL), which penalizes unnecessarily long trajectories by weighting SR with the ratio of the shortest path length to the actual trajectory length. For the more challenging REVERIE dataset, we further adopt two goal-specific metrics: Remote Grounding Success Rate (RGS), which evaluates whether the agent correctly localizes the target object bounding box; and Remote Grounding SPL (RGSPL), a weighted version of RGS that accounts for path efficiency.

\subsubsection{Implementation Details}

We use CLIP-B/16~\citep{radford2021learning} to extract image features and initialize our model with pretrained LXMERT weights~\citep{tan2019lxmert}. During the pretraining stage, both R2R and REVERIE are trained using five NVIDIA Tesla V100 GPUs with 16GB VRAM. We employ the AdamW optimizer~\citep{loshchilov2017decoupled} with a batch size of 6 per GPU and a learning rate of $1 \times 10^{-5}$. The total training steps are 355K for R2R and 100K for REVERIE.

During fine-tuning, we use a single NVIDIA A40 GPU, with a batch size of 16 and the learning rate fixed at $1 \times 10^{-5}$. Fine-tuning proceeds for 10K steps on R2R and 20K steps on REVERIE. The best models are selected based on SPL for R2R and RGSPL for REVERIE.

For pretraining tasks, The R2R dataset includes Single-step Action Prediction (SAP)~\citep{chen2021history}, Masked Language Modeling (MLM)~\citep{devlin2019bert}, and Masked Region Classification (MRC)~\citep{lu2019vilbert}. For REVERIE, we additionally include an Object Grounding (OG) task~\citep{lin2021scene}.

When building the text knowledge base, we generate captions for images with BLIP-2 in the LAVIS framework. We use the same prompt, “a photo of”, for all images to keep the captions consistent. All remaining settings follow the default BLIP-2 configuration in LAVIS.

For the image knowledge base, we generate images with Flux-Schnell in bfloat16. To match the indoor real-estate domain, we use the template prompt “[Goal Phrase] of the indoor environment in real estate”. We run sampling for 4 steps with guidance scale set to 0.0 and a maximum sequence length of 256. Images are generated at 1024 × 1024 resolution, and we fix the random seed to 0 for reproducibility.

\begin{table}[t]
\centering
\caption{Quantitative analysis of offline knowledge base construction costs.}
\label{tab:offline_cost}
\resizebox{\linewidth}{!}{%
\begin{tabular}{l c c c c c}
\toprule
\text{Component} & \text{Dataset} & \text{GPU} & \text{Time} & \text{Count} & \text{Storage} \\
\midrule
Textual KB & Matterport3D & 1$\times$ A40 & $\sim$120 h & 380K & 14 MB / 1.60 GB \\
\midrule
\multirow{2}{*}{Image KB} 
& R2R     & 2$\times$ A40 & $\sim$60 h & 93K & 11.57 GB / 189.24 MB \\
& REVERIE & 1$\times$ A40 & $\sim$63 h & 50K & 5.85 GB / 137.60 MB \\
\bottomrule
\end{tabular}%
}
\end{table}

\begin{table}[t]
\centering
\caption{Quantitative comparison of inference overhead between DUET and BTK.}
\label{tab:inference_cost}
\begin{tabular}{l c c c}
\toprule
\text{Method} 
& \text{Inference Time} 
& \text{VRAM Usage} 
& \text{Parameters} \\
& \text{(ms/step)} 
& \text{(GB)} 
& \text{(M)} \\
\midrule
DUET (Baseline) & 37 & 9  & 181 \\
BTK (Ours)     & 48 & 15 & 203 \\
\midrule
Overhead       & +11 & +6 & +22 \\
\bottomrule
\end{tabular}
\end{table}

\subsubsection{Efficiency Analysis} To assess the practical feasibility of BTK in real-world scenarios, we conduct a comprehensive quantification of the computational resources required, distinguishing between offline knowledge construction and online navigation inference.

\textbf{Offline Construction Costs}  \autoref{tab:offline_cost} reports the computational overhead of constructing the multimodal knowledge bases. It is important to emphasize that this overhead is a strictly one time preprocessing cost. For the text knowledge base, we use BLIP-2 to generate textual knowledge, which takes approximately 120 hours on a single NVIDIA A40 GPU, resulting in a corpus of about 380K entries with a raw size of roughly 14 MB. We then extract features using the CLIP text encoder, producing a feature bank of about 1.6 GB. For the image knowledge base, we adopt the Flux-Schnell model. On the R2R dataset, generating around 90K images takes about 60 hours using two A40 GPUs; on the REVERIE dataset, generating around 50K images takes about 63 hours on one A40 GPU. The raw generated images require approximately 11.6 GB (R2R) and 5.9 GB (REVERIE) of storage. Although storing the generated images is relatively costly, the navigation agent relies only on pre-extracted features at run time. After feature extraction with the CLIP image encoder, the resulting feature files are lightweight, namely 189 MB for R2R and 138 MB for REVERIE, which substantially reduces the storage footprint during inference.

\textbf{Inference Overhead}  \autoref{tab:inference_cost} provides a quantitative comparison of the online inference overhead between BTK and the DUET baseline on a single NVIDIA A40 GPU. By leveraging pre-computed feature banks and avoiding running generative models during inference, BTK remains efficient at run time. In terms of model size, DUET contains approximately 181M parameters, while BTK has about 203M parameters, an increase of roughly 22M. For online inference, the average per-step latency is 37 ms for DUET and 48 ms for BTK, corresponding to a modest increase of 11 ms. To accommodate the richer multimodal feature banks, VRAM consumption rises from 9 GB to 15 GB, i.e., an additional 6 GB. Despite the higher memory usage, the overall footprint remains below the capacity of standard high-end consumer and workstation GPUs, supporting the practicality of deploying BTK in real-world settings.

\begin{table*}[t]
\centering
\caption{Comparison with the state-of-the-art methods on the R2R dataset}
\label{tab:r2r}
\resizebox{1\textwidth}{!}{%
\begin{tabular}{lccccccccccccc}
\toprule
Methods & \multicolumn{4}{c}{Val Seen} & \multicolumn{4}{c}{Val Unseen} & \multicolumn{4}{c}{Test Unseen} \\
\cmidrule(lr){2-5} \cmidrule(lr){6-9} \cmidrule(lr){10-13}
 & NE$\downarrow$ & OSR$\uparrow$ & SR$\uparrow$ & SPL$\uparrow$
 & NE$\downarrow$ & OSR$\uparrow$ & SR$\uparrow$ & SPL$\uparrow$
 & NE$\downarrow$ & OSR$\uparrow$ & SR$\uparrow$ & SPL$\uparrow$ \\ 
\midrule
Seq2Seq\citep{anderson2018vision} & 6.01 & - & 39 & - & 7.81 & - & 22 & - & 7.85 & - & 20 & 18 \\ 
EnvDrop \citep{tan2019learning} & 3.99 & - & 62 & 59 & 5.22 & - & 52 & 48 & 5.23 & 59 & 51 & 47 \\ 

HOP+ \citep{qiao2023hop+} & 2.33 & - & 78 & 73 & 3.49 & - & 67 & 61 & 3.71 & - & 66 & 60 \\ 
HAMT \citep{chen2021history} & 2.34 & 82 & 76 & 72 & 2.29 & 73 & 66 & 61 & 3.93 & 72 & 65 & 60 \\ 
DUET \citep{chen2022think} & 2.28 & 86 & 79 & 73 & 3.31 & 81 & 72 & 60 & 3.65 & 76 & 69 & 59 \\ 
TD-STP \citep{zhao2022target} & 2.34 & 83 & 77 & 73 & 3.22 & 76 & 70 & 63 & 3.73 & 72 & 67 & 61 \\ 
GridMM \citep{wang2023gridmm} & 2.34 & - & 80 & 74 & 2.83 & - & \textbf{75} & \textbf{64} & 3.35 & - & 73 & 62 \\ 
LSAL \citep{wu2024vision} & 2.88 & - & 73 & 70 & 3.62 & - & 65 & 59 & 4.00 & - & 63 & 58 \\ 
NavGPT2 \citep{zhou2024navgpt} & 2.84 & 83 & 74 & 63 & 2.98 & \textbf{84}  & 74 & 61 & 3.33 & 80 & 72 & 60 \\ 
KERM \citep{li2023kerm} & 2.19 & - & 80 & 74 & 3.22 & - & 72 & 61 & 3.61 & - & 70 & 59 \\ 
KESU\citep{gao2024enhancing} & 2.19 & - & 81 & 75 & 2.96 & - & 73 & 62 & 3.31 & - & 72 & 61 \\ 
BEVBert \citep{an2023bevbert} & 2.17 & - & 81 & 74 & 2.81 & - & \textbf{75} & \textbf{64} & 3.13 & \textbf{81} & 73 & 62 \\ 
BSG \citep{liu2023bird} & - & - & - & - & 2.80 & - & 74 & 62 & 3.19 & - & 73 & 62 \\ 
CONSOLE \citep{lin2024correctable} & 2.17 & - & 79 & 73 & 3.00 & - & 73 & 63 & 3.30 & - & 72 & 61 \\ 
Esceme \citep{zheng2025esceme} & 2.57 & - & 76 & 73 & 3.39 & - & 68 & \textbf{64}& 3.77 & - & 66 & \textbf{63} \\ 
TSOR \citep{huang2025temporal} & \textbf{2.05} & - & \textbf{82} & \textbf{76} & 3.05 & - & 71 & 61& 3.27 & - & 72 & 62 \\ 
ViTeC \citep{gao2025visual} & 2.22 & - & 79 & 74 & 3.15 & - & 72 & 62& 3.31 & - & 70 & 60 \\ 

\textbf{BTK(Ours)} & 2.09 & \textbf{88} & 81 & 75 & \textbf{2.69} & \textbf{84} &  \textbf{75} & \textbf{64} & \textbf{3.07} & \textbf{81} & \textbf{74} & \textbf{63} \\
\bottomrule
\end{tabular}
}
\end{table*}

\begin{table*}[t]
\centering
\caption{Comparison with the state-of-the-art methods on the REVERIE dataset}
\label{tab:reverie}
\resizebox{1\textwidth}{!}{%
\begin{tabular}{lcccccccccc}
\toprule
Methods& \multicolumn{5}{c}{Val Unseen} & \multicolumn{5}{c}{Test Unseen} \\
\cmidrule(lr){2-6} \cmidrule(lr){7-11}
& OSR$\uparrow$ & SR$\uparrow$ & SPL$\uparrow$ & RGS$\uparrow$ & RGSPL$\uparrow$ 
& OSR$\uparrow$ & SR$\uparrow$ & SPL$\uparrow$ & RGS$\uparrow$ & RGSPL$\uparrow$ \\
\midrule
Airbert\citep{guhur2021airbert} & 34.51 & 27.89 & 21.88 & 18.23 & 14.18 & 34.20 & 30.28 & 23.61 & 16.83 & 13.28 \\ 
HOP+\citep{qiao2023hop+} & 40.04 & 36.07 & 31.13 & 22.49 & 19.33 & 35.81 & 33.82 & 28.24 & 20.20 & 16.86 \\ 
HAMT\citep{chen2021history} & 36.84 & 32.95 & 30.20 & 18.92 & 17.28 & 33.41 & 30.40 & 26.67 & 14.88 & 13.08 \\ 
TD-STP\citep{zhao2022target} & 39.48 & 34.88 & 27.32 & 21.16 & 16.56 & 40.26 & 35.89 & 27.51 & 19.88 & 15.40 \\ 
DUET\citep{chen2022think} & 51.07 & 46.98 & 33.73 & 32.15 & 23.03 & 56.91 & 52.51 & 36.06 & 31.88 & 22.06 \\ 
LANA \citep{wang2023lana} & 52.97 & 48.31 & 33.86 & 32.86 & 22.77 & 57.20 & 51.72 & 36.45 & 32.95 & 22.85 \\ 
FDA\citep{he2023frequency} & 51.41 & 47.57 & 35.90 & 32.06 & 24.31 & 53.54 & 49.62 & 36.45 & 30.34 & 22.08 \\ 
KERM\citep{li2023kerm} & 55.21 & 49.02 & 34.83 & 33.97 & 24.14 & 57.44 & 52.26 & 37.46 & 32.69 & 23.15 \\ 
BEVBert\citep{an2023bevbert} & 56.40 & 51.78 & 36.37 & 34.71 & 24.44 & 57.26 & 52.81 & 36.41 & 32.06 & 22.09 \\ 
VER\citep{liu2024volumetric} & \textbf{61.09} & \textbf{55.98} & \textbf{39.66} & 33.71 & 23.70 & \textbf{62.22} & 56.82 & 38.76 & 33.88 & 23.19 \\ 
ACME\citep{wu2025adaptive} & 53.97 & 49.46 & 32.37 & 32.64 & 24.02 & 57.48 & 51.89 & 34.65 & 33.12 & 23.57 \\ 
RBCRN\citep{wu2025recursive} & 52.71 & 48.00 & 35.14 & 32.72 & 24.16 & 56.56 & 51.38 & 33.75 & 32.26 & 22.63 \\
ACK\citep{mohammadi2024augmented} & 52.77 & 47.49 & 34.44 & 32.66 & 23.92 & 59.01 & 53.97 & 37.89 & 32.77 & 23.15 \\ 
ViTeC \citep{gao2025visual}& 54.47 & 50.18 & 35.06 & \textbf{34.82} & 24.23 & 62.16 & \textbf{57.52} & 38.09 & 34.09 & 22.81 \\ 

% \midrule
\textbf{BTK(Ours)} & 56.80 & 50.47 & 35.88 & \textbf{34.82} & \textbf{24.86} & 61.16 & 54.58 & \textbf{39.75}  & \textbf{35.08} & \textbf{25.23} \\
\bottomrule
\end{tabular}
}
\end{table*}

\subsection{Quantitative Results}
\subsubsection{Comparison with Existing Methods}

To comprehensively evaluate the proposed method, we conduct experiments on the R2R~\citep{anderson2018vision} and REVERIE~\citep{qi2020reverie} datasets and compare our model with several state-of-the-art baselines. The quantitative results are summarized in  \autoref{tab:r2r} and  \autoref{tab:reverie}.

On the R2R dataset, our method significantly outperforms the strong baseline DUET across multiple key metrics. Specifically, on the seen validation set, our model achieves SR and SPL scores of 81 and 75, respectively, both representing 2\% improvements over DUET. On the unseen validation set, SR and SPL reach 75 and 64, marking 3\% and 4\% improvements, respectively.

On the more challenging REVERIE dataset, our method also demonstrates strong performance. Specifically, on the unseen validation split, we achieve improvements of 4.25\%, 2.07\%, 3.69\%, 3.20\%, and 3.17\% over the baseline model in the OSR, SR, SPL, RGS, and RGSPL metrics, respectively. Compared to the current state-of-the-art model ACK, our approach attains gains of 2.31\% and 2.08\% in the key metrics RGS and RGSPL, respectively, highlighting the significant advantage of our constructed multimodal knowledge base in enhancing goal-awareness.

\subsection{Ablation Studies}
We conduct comprehensive ablation studies on the REVERIE val unseen set to validate the effectiveness of each component in our proposed model.
\begin{table}[ht]
\centering
\caption{Ablation Study Comparing Different Extraction Methods}
\label{tab:gaa}
% \scriptsize % 可根据需要调整字体大小,如 \footnotesize 或 \scriptsize
\begin{tabular}{cccccc}
\toprule
\text{Method} & \text{SR↑} & \text{SPL↑} & \text{RGS↑} & \text{RGSPL↑} \\ 
\midrule
\text{Baseline}  & 46.98 & 33.73 & 32.15 & 23.03 \\ 
\text{+ Spacy}  & 47.94 & \textbf{33.77} & 33.12 & \textbf{23.60} \\  
\text{+ Qwen3-4B}  & \textbf{50.03} & 33.32 & \textbf{34.05} & 23.35 \\ 
\bottomrule
\end{tabular}
\end{table}

\begin{table}[h]
\centering
\caption{Ablation Study on Multimodal Knowledge Integration}
\label{tab:know7}
\begin{tabular}{cccccccc} 
\toprule
\text{Id} & \text{IK}  & \text{TK}  & \text{SR$\uparrow$}  & \text{OSR$\uparrow$} & \text{SPL$\uparrow$ } & \text{RGS$\uparrow$}  & \text{RGSPL$\uparrow$}  \\
\midrule
\text{1}  & $\times$ & $\times$& 50.30& 55.97 & 33.32 & 34.05 & 23.35 \\ 
\text{2} & $\checkmark$ & $\times$& 49.59& 56.18 & 32.68 & 34.76 & 23.27 \\ 
\text{3} & $\times$ & $\checkmark$ & 48.88& 55.33 & 33.18 & 34.11 & 23.84 \\ 
\text{4} & $\checkmark$ & $\checkmark$ & \textbf{50.47} & \textbf{56.80} & \textbf{35.88} & \textbf{34.82} & \textbf{24.86} \\ 
\bottomrule
\end{tabular}
\end{table}

\begin{table}[h]
\centering
\caption{Ablation Study on Key Components of Multimodal Knowledge Integration, $\sigma$ denotes the sigmoid function.}
\label{tab:know}
\begin{tabular}{cccccccc} 
\toprule
\text{} & \text{IK}  & \text{TK}  & \text{SR$\uparrow$}  & \text{OSR$\uparrow$} & \text{SPL$\uparrow$ } & \text{RGS$\uparrow$}  & \text{RGSPL$\uparrow$}  \\
\midrule
$\sigma$  & $w/o$ & $w/o$& 50.04& 56.66 & 33.47 & 34.37 & 23.24 \\ 
$\sigma$ & $w/$ & $w/o$& 49.93& 55.55 & 34.33 & 33.66 & 23.37 \\ 
$\sigma$ & $w/o$ & $w/$ & 48.71& 54.96 & 34.19 & 33.20 & 23.90 \\ 
$\sigma$ & $w/$& $w/$ & \textbf{50.47} & \textbf{56.80} & \textbf{35.88} & \textbf{34.82} & \textbf{24.86} \\ 
\bottomrule
\end{tabular}
\end{table}

\begin{table}[h]
\centering
\caption{Ablation Study on the Role of Goal-Aware and Knowledge Augmentors}
\label{tab:aug}
\begin{tabular}{cccccccc} 
\toprule
\text{Id}  & \text{GAA}  & \text{KA}  & \text{SR$\uparrow$}  & \text{OSR$\uparrow$} & \text{SPL$\uparrow$}  & \text{RGS$\uparrow$}  & \text{RGSPL$\uparrow$}  \\
\midrule
1  & $\times$ & $\times$& 48.76& 55.50 & 32.87 & 34.48 & 23.22 \\ 
2  & $\checkmark$ & $\times$& 46.80& 52.63 & 33.39 & 32.49 & 23.46 \\ 
3  & $\times$ & $\checkmark$ & 48.91& 56.26 & 32.85 & 34.05 & 23.38 \\ 
4  & $\checkmark$ & $\checkmark$ & \textbf{50.47} & \textbf{56.80} & \textbf{35.88} & \textbf{34.82} & \textbf{24.86} \\ 
\bottomrule
\end{tabular}
\end{table}

\begin{table}[t]
    \centering
    \caption{Quantitative analysis of feature representation changes.}
    \label{tab:feature_analysis}
    \begin{tabular}{llccc}
        \toprule
        \text{Modality} & \text{Metric} & \text{Before} & \text{After} & \text{Change (\%)} \\
        \midrule
        \multirow{2}{*}{Visual} & Variance & 0.363 & 0.295 & $-$18.69 \\
                                & Avg Dist & 23.195 & 20.930 & $-$9.76 \\
        \midrule
        \multirow{2}{*}{Text}   & Variance & 0.098 & 0.146 & +49.97 \\
                                & Avg Dist & 12.074 & 14.818 & +22.73 \\
        \bottomrule
    \end{tabular}
\end{table}

\subsubsection{Target-Aware Augmentation Module}

\begin{figure*}[!t]
    \centering
    \includegraphics[width=\textwidth]{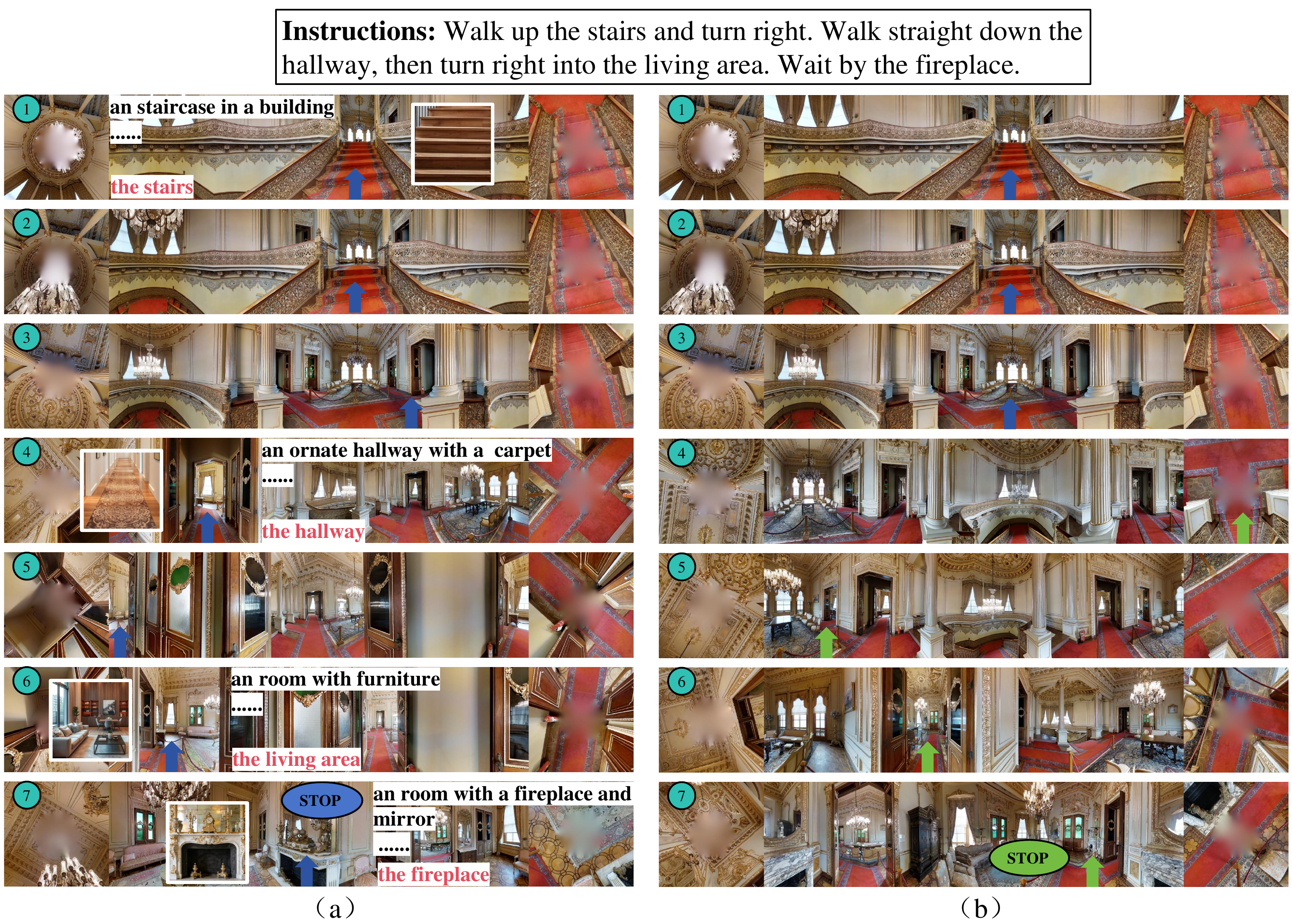}
    \caption{Trajectory visualizations on R2R val seen. (a) Ground truth and BTK prediction. (b) DUET prediction.}
    \label{rjg}
\end{figure*}

\begin{figure*}[h]
    \centering
    \includegraphics[width=\textwidth]{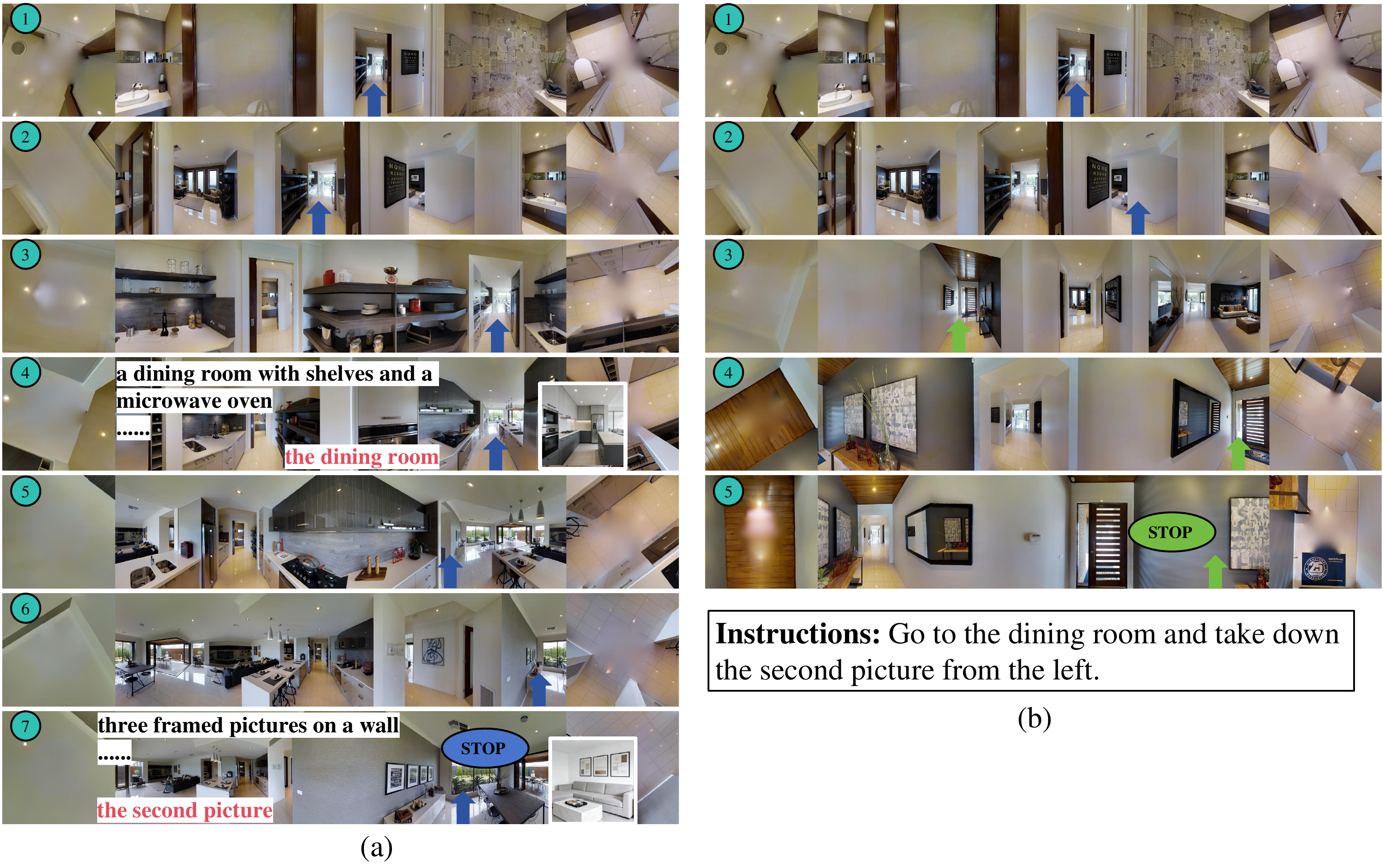}
    \caption{Trajectory visualizations on REVERIE val unseen. (a) Ground truth and BTK prediction. (b) DUET prediction.}
    \label{rwj}
\end{figure*}

\begin{figure*}[h]
    \centering
    \includegraphics[width=\textwidth]{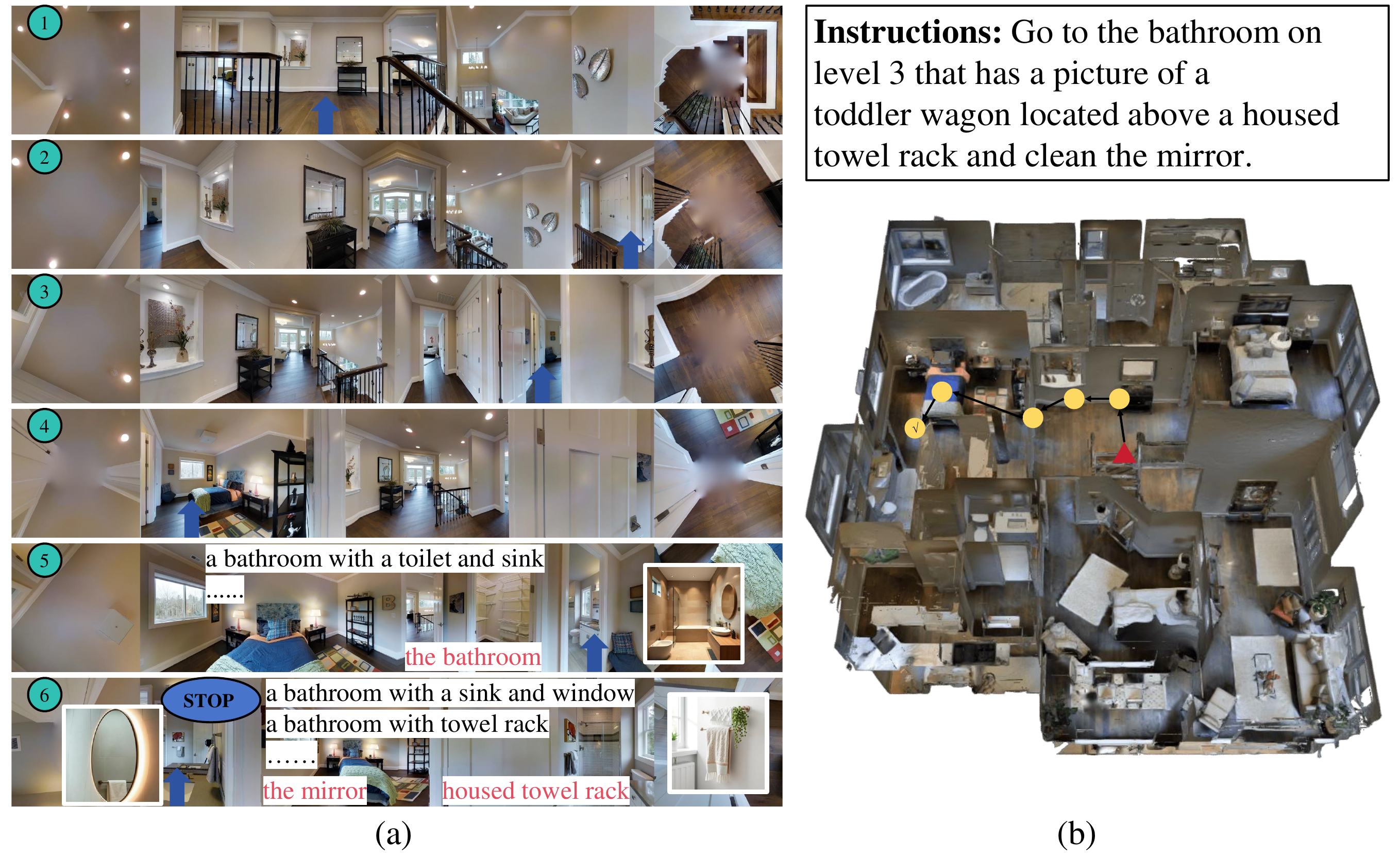}
    \caption{BTK visualization on REVERIE val unseen. (a) Trajectory visualization. (b) Top-down view of navigation path.}
    \label{fs}
\end{figure*}

\begin{figure*}[h]
    \centering
    \includegraphics[width=\textwidth]{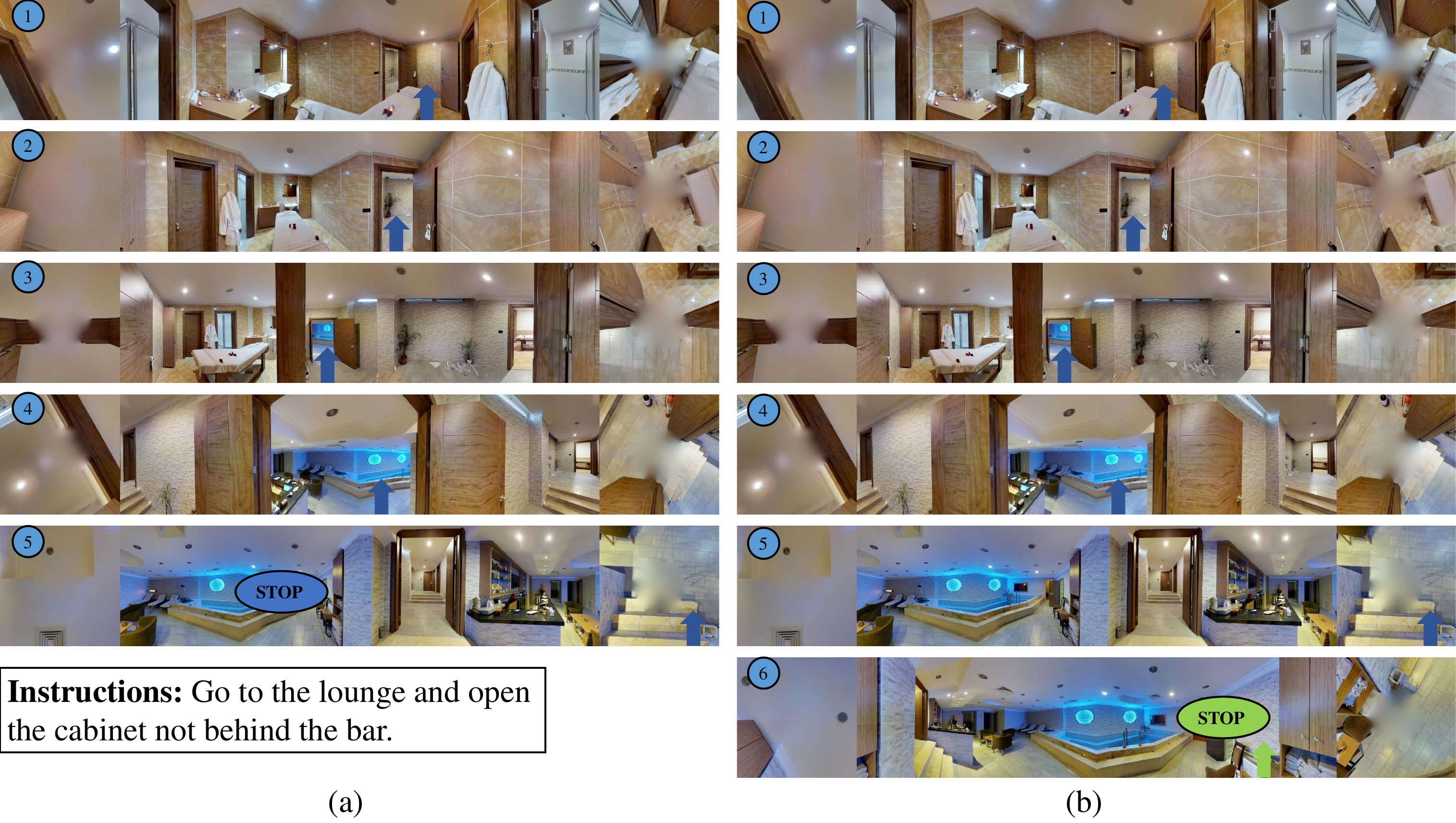}
    \caption{Failure case on REVERIE val unseen. (a) Ground-truth trajectory. (b) BTK prediction.}
    \label{cw}
\end{figure*}

\begin{figure*}[h]
    \centering
    \includegraphics[width=\textwidth]{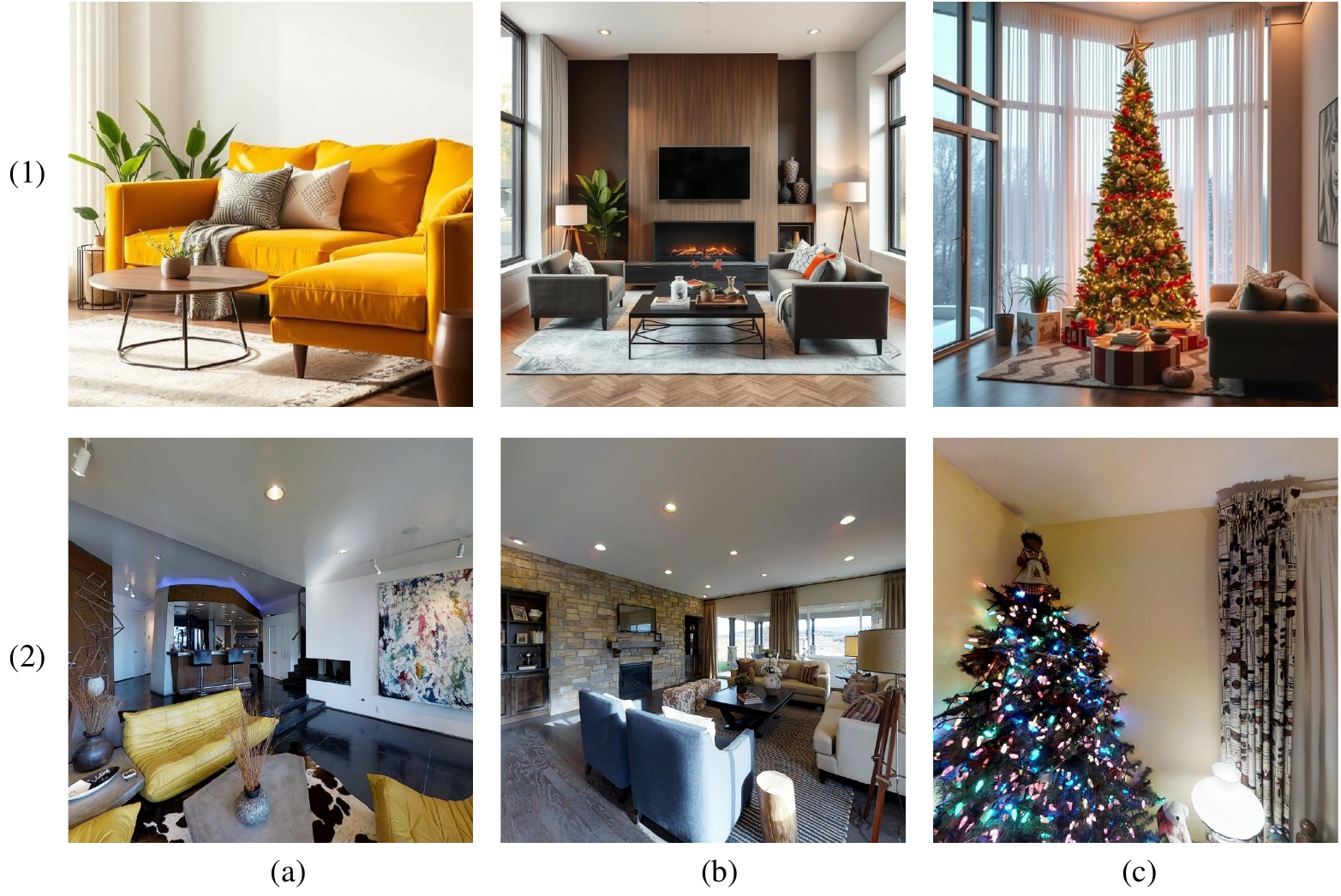}
    \caption{The differences between the images generated by Flux-Schnell and the real images are shown in the figure. Columns (a)–(c) respectively present the generated results for the yellow lounge sofa, the living room, and the Christmas tree; the first row shows the generated images, while the second row shows the corresponding real images.}
    \label{cj}
\end{figure*}
To evaluate the impact of different target phrase extraction methods, we compare using SpaCy and Qwen3-4B for extracting subgoal phrases from instructions. As shown in  \autoref{tab:gaa}, while SpaCy offers slight improvements over the baseline, using Qwen3-4B yields significant gains, improving SR and RGS by 3.05\% and 1.9\%, respectively. This demonstrates Qwen3-4B's superior semantic understanding in target phrase extraction, which directly benefits navigation performance.

\subsubsection{Multimodal Knowledge Ablation}

To examine the impact of multimodal knowledge on navigation tasks, we integrate textual knowledge (TK), image knowledge (IK), and their combination into the DUET model equipped with goal-aware augmentation. The results are reported in \autoref{tab:know7}. We observe a slight drop in SPL when only image knowledge is used, which is mainly attributed to discrepancies between generated images and real-world environments in terms of visual fidelity and feature representations, potentially undermining execution stability. In contrast, incorporating the full multimodal knowledge base yields consistent and significant gains across all metrics, outperforming the settings that use only IK or only TK. These findings suggest that multimodal knowledge provides strong complementarity in visual grounding and semantic completion, thereby enhancing the model’s understanding and decision-making in complex navigation environments.

\subsubsection{Ablation of Key Sub-modules for Multimodal Knowledge}

 \autoref{tab:know} analyzes the role of sigmoid gating in multimodal knowledge injection when both image knowledge (IK) and textual knowledge (TK) are available. Without gating, enhanced features are directly adopted, resulting in hard and uncontrolled knowledge injection, while sigmoid gating dynamically fuses enhanced and original features. Disabling gating on either pathway degrades performance, especially SPL, whereas enabling gating on both yields the best and most stable results.

\subsubsection{Augmentor Ablation}

To assess the independent and joint effects of the Goal-Aware Augmentor (GAA) and Knowledge Augmentor (KA), we evaluate several model variants on the val unseen set.  \autoref{tab:aug} shows that the full model using both GAA and KA achieves the best performance across all metrics, confirming their structural complementarity and the synergy between goal comprehension and knowledge utilization.

\subsection{Analysis of Feature Representation Changes}
To explore the intrinsic mechanism of the Knowledge Augmentor, we conducted feature separability analysis on textual and visual features respectively, with the results presented in \autoref{tab:feature_analysis}.

\subsubsection{Visual Modality} We observe that the mean pairwise distance of visual features decreases from 23.195 to 20.930, a reduction of 9.76\%, and the variance drops from 0.363 to 0.295, a reduction of 18.69\%. In complex indoor scenes, raw visual observations are often influenced by background noise, such as ceilings and cluttered furnishings, leading to a more dispersed distribution in the feature space. The simultaneous decreases in these two metrics indicate that our method effectively suppresses task-irrelevant interference, making the feature distribution more compact and enabling the representation to focus more stably on regions that are relevant to navigation decisions.

\subsubsection{Text Modality} In contrast, for the text modality, the mean pairwise distance increases from 12.074 to 14.818 (a 22.73\% increase), and the variance rises from 0.098 to 0.146 (a 49.97\% increase). Standard navigation instructions are often semantically sparse, which is reflected by low variance in the embedding space. By injecting image knowledge into the textual instruction, our method broadens the underlying semantic manifold and increases the diversity of textual representations. The larger pairwise distances suggest improved separability, indicating that the enhanced instructions become more distinctive from one another.

\subsection{Qualitative Results and Visualization}

To provide a more intuitive demonstration of BTK’s navigation performance, we visualize its navigation trajectories on the R2R and REVERIE datasets, as shown in \autoref{rjg} to \ref{fs}. In these figures, blue arrows indicate correct navigation paths, while green arrows indicate incorrect paths. Text shown in red on a white background denotes the goal phrase, text shown in black on a white background denotes the textual knowledge, and the images enclosed by white boxes correspond to the image knowledge associated with the goal phrase. Additionally, to compare the differences between images generated by Flux-Schnell and real images, we present a comparison in \autoref{cj}.

 \autoref{rjg} presents a comparison of navigation trajectories between our BTK model and the baseline DUET on the R2R val seen subset.  \autoref{rjg}(a) shows BTK’s predicted trajectory, which matches the ground-truth path at every step, demonstrating the model’s strong capability in instruction understanding and visual alignment. In contrast,  \autoref{rjg}(b) depicts DUET’s trajectory, which deviates noticeably at step four and ultimately fails to reach the correct goal location, mainly due to the lack of goal-awareness enhancement and multimodal knowledge guidance.

 \autoref{rwj} illustrates navigation trajectory examples of BTK and the baseline DUET on the REVERIE val unseen split. In this example, the model receives a brief navigation instruction describing only the goal location, unlike the detailed instructions in R2R, making the task more challenging.  \autoref{rwj}(a) shows BTK’s navigation trajectory prediction process, where a total of seven action steps are taken from start to finish, perfectly matching the ground-truth path. Our model aligns textual knowledge obtained from the knowledge base with complementary visual information, while also incorporating image knowledge for alignment, enabling accurate navigation even under brief instruction conditions.  \autoref{rwj}(b) shows DUET’s predicted trajectory, which makes an incorrect decision at step three due to the absence of multimodal knowledge, ultimately stopping at an incorrect location.

To provide a clearer and more objective demonstration of BTK’s decision-making process, we present a top-down navigation trajectory visualization of the agent on the REVERIE val unseen split.  \autoref{fs}(b) shows a floor plan of the entire room, where the red triangle marks the starting position and the yellow circle represents the navigation path predicted by our model.From the bird’s-eye view of the overall trajectory, BTK accurately identifies the forward direction and ultimately stops at the target location.

\subsection{Failure Case Visualization}

Despite strong overall performance, BTK may still fail in complex scenarios.  \autoref{cw} shows a representative failure case.

In \autoref{cw} (a), the ground truth trajectory includes five action steps. BTK's prediction ( \autoref{cw}(b)) follows the correct path until the final step, when it incorrectly performs one additional move, failing to stop at the goal. This indicates a need for improved termination strategies, suggesting future work should focus on modeling the stopping condition more robustly.

\subsection{Generated–Real Image Comparison}
To intuitively demonstrate the effectiveness of the generated images, we selected three types of examples for comparison: images corresponding to descriptive phrases, room-related phrases, and specific objects. As shown in  \autoref{cj}, regardless of the type, the images generated by Flux-Schnell exhibit a high degree of similarity to the real images.

\section{Implications}
\subsection{Theoretical Implications}
Our contribution to the field of Embodied AI lies in shifting the paradigm of knowledge retrieval from external text-only sources to multimodal knowledge integration. Unlike prior work that relies on structured external text knowledge graphs\citep{li2023kerm,mohammadi2024augmented}
, the BTK framework demonstrates that internal textual knowledge can provide semantic cues for vision, and that generative visual knowledge can serve as a more direct bridge for cross-modal alignment. By converting abstract textual goals into concrete visual “hallucinations” (via R2R\_GP and REVERIE\_GP), we provide a theoretical foundation for using generative AI as a “visual interpreter” for navigation agents, thereby reducing the cognitive load of mapping language to perception.
\subsection{Practical Implications}
At the application level, the BTK framework effectively mitigates the critical failure mode of “stopping at the wrong location” caused by ambiguous target-object descriptions. The Goal-Aware Augmentor parses complex instructions more effectively than standard noun-extraction methods. For real-world deployment, this implies that service robots can invoke efficient local generative models to “imagine” the target before acting, thereby improving robustness to natural, unconstrained user commands.
\section{Conclusion}
We present BTK, a novel VLN framework that integrates goal-awareness and multimodal knowledge to address limitations in language understanding and visual grounding. By leveraging Qwen3-4B to extract target phrases and enhancing goal comprehension, BTK aligns key instruction semantics with visual cues via textual and image knowledge bases. Experimental results on the R2R and REVERIE datasets demonstrate that BTK surpasses existing state-of-the-art methods on multiple evaluation metrics, showcasing its strong navigation performance.

\section{Limitations}
Our framework relies on pretrained generative models: Qwen3-4B is used for target phrase extraction, Flux-Schnell is adopted to construct image knowledge, and BLIP-2 is employed to build textual knowledge. While these models are generally stable on common indoor objects and attributes, their generation quality may degrade when instructions involve rare targets, fine-grained attributes, or novel compositional concepts, which are under-represented in the training distribution. Such degradation can weaken or reduce the precision of the semantic cues provided by the image or textual knowledge.

It is important to emphasize that BTK does not directly use the generated image or textual knowledge for perception or action decision-making. Instead, the generated content is encoded into CLIP features and then dynamically fused with the original features through a gated attention mechanism, balancing the original representations and the knowledge-enhanced ones. This design ensures that generation errors mainly appear as attenuated auxiliary alignment signals rather than misleading the navigation policy. Therefore, when facing rare targets or novel object compositions, performance typically degrades in a controlled manner, and the agent can fall back on the original instruction representations and real-world environmental observations to make decisions.

\section{Acknowledgement}
This research was financially supported by the National Natural Science Foundation of China (Grant No. 62463029).

\bibliographystyle{elsarticle-harv} 
\bibliography{ref.bib}
\end{document}